\documentclass{article}
\usepackage[margin=2.5cm]{geometry}
\usepackage{hyperref}
\usepackage[round]{natbib}
\usepackage{amsthm}

\usepackage{amssymb}
\usepackage{epstopdf}

\usepackage{subfigure}
\usepackage{mathtools}
\usepackage{amsfonts}
\usepackage{amsmath}
\usepackage{graphicx}
 \usepackage{booktabs}
\usepackage{algorithm}
\usepackage[table,xcdraw]{xcolor}
\usepackage{algpseudocode}

\newtheorem{lemma}{Lemma}

\newcommand{\OMIT}[1]{}
\newcommand{\RR}{\mathbb{R}}    %reals
\newcommand {\br}[1]{\left(#1\right)}
\newcommand{\Eb}{\mathbb{E}}    %expectation \E(x)
\newcommand{\One}{\mathbf{1}}   
\newcommand{\Id}{\mathbf{I}}   
\newcommand{\Dcal}{\mathcal{D}}

\newcommand{\beginsupplement}{%
        \setcounter{table}{0}
        \renewcommand{\thetable}{S\arabic{table}}%
        \setcounter{figure}{0}
        \renewcommand{\thefigure}{S\arabic{figure}}%
     }

\begin{document}

\title{Adaptive structured noise injection for shallow and deep neural networks}
\author{Beyrem Khalfaoui\,$^{\text{1,2}}$ , Joseph Boyd\,$^{\text{1,2}}$, Jean-Philippe Vert\,$^{\text{3,1}}$\\ \\
$^{\text{\sf 1}}$  MINES ParisTech, PSL Research University, \\CBIO - Centre for Computational Biology, F-75006 Paris, France\\
$^{\text{\sf 2}}$ Institut Curie, PSL Research University, INSERM, U900, F-75005 Paris, France.\\    
$^{\text{\sf 3}}$ Google Brain, F-75009 Paris, France.\\
\url{beyrem.khalfaoui@mines-paristech.fr}, \url{jpvert@google.com}}
\date{}
\maketitle{}

\begin{abstract}
\noindent Dropout is a regularisation technique in neural network training where unit activations are randomly set to zero with a given probability \emph{independently}. In this work, we propose a generalisation of dropout and other multiplicative noise injection schemes for shallow and deep neural networks, where the random noise applied to different units is not independent but follows a joint distribution that is either fixed or estimated during training. We provide theoretical insights on why such adaptive structured noise injection (ASNI) may be relevant, and empirically confirm that it helps boost the accuracy of simple feedforward and convolutional neural networks, disentangles the hidden layer representations, and leads to sparser representations.
Our proposed method is a straightforward modification of the classical dropout and does not require additional computational overhead. \\
%\textbf{Contact:} \href{jean-philippe.vert@mines-paristech.fr}{jean-philippe.vert@mines-paristech.fr}\\
%\textbf{Supplementary information:} Supplementary data are available at \textit{Bioinformatics} online.}
\end{abstract}

\section{Introduction}
\label{sec:intro}

The tremendous empirical success of deep neural networks (DNN) for many machine learning tasks such as image classification and object recognition~\citep{krizhevsky_imagenet_2017} contrasts with their relatively poor theoretical understanding. One feature commonly attributed to DNN to explain their performance is their ability to build hierarchical \emph{representations} of the data, able to capture relevant information in the data at different scales~\citep{bengio_representation_2012,tishby_deep_2015,Mallat2012Group}. An important idea to create good sets of representations is to reduce redundancy and increase diversity in the representation, an idea that can be traced back to early investigations about learning~\citep{barlow1959possible} and that has been implemented in a variety of methods such as independent component analysis~\citep{hyvarinen_independent_2013} or feature selection~\citep{peng2005feature}. Explicitly encouraging diversity has been shown to improve the performance of ensemble learning models~\citep{kuncheva2003measures,dietterich2000ensemble}, and techniques have been proposed to limit redundancy in DNN by pruning units or connections~\citep{hassibi1993second,lecun1990optimal,mariet2016diversity} or by explicitly encouraging diversity between units of each layer during training~\citep{cogswell2015reducing,desjardins2015natural,rodriguez2016regularizing,luo2017learning}. 

Dropout~\citep{hinton_improving_2012,srivastava2014dropout} is a recent and popular regularisation techniques in deep learning that exploits the idea of creating diversity stochastically while training, by randomly setting to zero come units or connections during stochastic gradient optimization. Proposed by~\citet{hinton_improving_2012} as a way to prevent co-adaptation of units and to approximately combine exponentially many different DNN architectures efficiently, it has improved DNN performance in many benchmark datasets. Dropout can also be interpreted as a regularization technique \citep{baldi2013understanding,wager2013dropout,maeda_bayesian_2014,helmbold2017surprising}, however its impact on learning a good representation of the data remains elusive. Several variants of the original dropout model have been proposed to adapt the algorithm to other models~\citep[e.g.,][]{gal2016theoretically}, or to modify the distribution of the stochastic noise. \citet{ba2013adaptive} propose that the dropout rate of a unit should depend on its magnitude and dynamically adapt the dropout rate of each unit activation during training. Another variant is to group units together because of their proximity in a map~\citep{tompson_efficient_2014,devries2017improved} of because they are strongly correlated~\citep{aydore_using_2018}, before applying dropout jointly on the units in a group. This can equivalently be interpreted as applying dropout to individual units, but constraining the stochastic dropout noise in units within a group to be perfectly correlated.

In this work, we extend and generalize the idea to modify the noise distribution, and analyse both theoretically and empirically the effect of different choices. In particular, we study the impact of creating correlations among noise in the units of a given layer, generalizing the ideas of \citet{tompson_efficient_2014,devries2017improved,aydore_using_2018} to a general covariance structure. We depart from binary dropout noise to the more flexible multiplicative Gaussian noise model, which allows to specify any covariance structure without the need to explicitly cluster units into groups, and highlight the role of the noise correlation matrix in the regularization effect of this structured noise injection procedure. We show in particular that borrowing the covariance of the units to create the covariance of the noise can decrease redundancy among the units, a phenomenon we confirm empirically that leads to better representations and classification accuracy.

\section{Dropout and multiplicative noise}		
Let us first set notations to describe a standard neural network for inputs in $\RR^d$ with $H$ layers of respective dimensions $d_1,\ldots,d_H$. For any layer $l\in [1,H]$ let $W^{(l)} \in \RR^{d^{(l)}\times d^{(l-1)}}$ and $b^{(l)} \in \RR^{d^{(l)}}$ denote respectively the matrix of weights and the vector of biases at layer $l$, with the convention $d^{(0)}=d$, and let $\theta = \br{W_l,b_l}_{l=1,\ldots,H}$ denote the set of parameters of the network. The network defines a function $f_\theta : \RR^d \rightarrow \RR^{d^{(H)}}$ given for any $x\in\RR^d$ by $f_\theta(x)=y^{(H)}$, where $y^{(l)}$ is defined recursively for $l=0,\ldots,H$ by $y^{(0)}=x$ and, for $l\in[1,H]$:
$$
\begin{cases}
z^{(l)} &=W^{(l)} y^{(l-1)}+ b^{(l)}\,,  \\
y^{(l)} &= \sigma^{(l)}(z^{(l)})\,,  
\end{cases}
$$
where $\sigma^{(l)}$ is an activation function at the $l$-th layer, such as the RELU function $\sigma(t)=t {\bf 1}(t>0)$ for $t\in\RR$, applied entrywise if its input is a vector.

Given a training set $\Dcal$ of $N$ labelled inputs $(x_1, y_1), \ldots, (x_N,y_N) \in \RR^d\times \RR^p$, and a loss function $L: \RR^{d^{(H)}} \times \RR^p \rightarrow \RR$, training the neural network amounts to fitting the parameters $\theta$ by trying to minimize the average loss over the training set:
\begin{equation}\label{eq:erm} 
\min_\theta \frac{1}{N}  \sum_{i=1}^{N}  L \left( f_\theta\br{x_i} , y_{i} \right )\,,
\end{equation} 
usually by some form of stochastic gradient descent (SGD) using backpropagation to compute gradients. For example, when the label is a scalar ($p=1$), then one can use $d^{(H)}=1$ and the squared error $L(u,v)=(u-v)^2$ for $u,v\in\RR$.

A popular way to improve the training of neural networks is to use dropout regularisation, where during training the units of the input and hidden layers are stochastically omitted during the back-propagation algorithm~\citep{srivastava2014dropout}. Dropout is a particular case of multiplicative noise, which we can formalize as follows. Given a sequence of vectors $r=\br{r^{(0)}, \ldots, r^{(H-1)}}\in\RR^{d^{(0)}}\times\ldots\times \RR^{d^{(H-1)}}$, of total dimensions $D=d^{(0)}+ \ldots+ d^{(H-1)}$, we create the modified function $f_\theta(x,r) = y^{(H)}$ where $y^{(0)}=x$ and, for $l\in[1,H]$:
$$
\begin{cases}
\widetilde{y}^{(l-1)} &= r^{(l-1)} \odot y^{(l-1)}\,,\\
z^{(l)} &=W^{(l)} \widetilde{y}^{(l-1)}+ b^{(l)}\,,  \\
y^{(l)} &= \sigma^{(l)}(z^{(l)})\,,  
\end{cases}
$$
where $\odot$ denotes the Hadamard or element-wise product. We then define a set of independent and identically distributed (i.i.d.) random variables $(R_i)_{i=1,\ldots,N}$ with values in $\RR^{D}$ (the ``noise'') and train the network by solving
\begin{equation}\label{eq:asni}
\min_\theta \frac{1}{N}  \sum_{i=1}^{N}  \Eb L \left( f_\theta\br{x_i, R_i} , y_{i} \right )\,.
\end{equation} 
With these notations, the standard dropout approach of~\citep{srivastava2014dropout} with parameter $p\in[0,1]$ is obtained by taking a noise distribution with i.i.d. entries across the dimensions taking the value $1/p$ with probability $p$, and $0$ with probability $1-p$.

\section{Structured noise injection (SNI)}
We propose to create new learning schemes by learning with multiplicative noise, as described above, where the noise distribution of $R$ is \emph{not} i.i.d across the units (while we keep the noise samples $R_1,\ldots,R_N$ i.i.d. according to the noise distribution). For simplicity, we focus only on Gaussian noise:
\begin{equation}\label{eq:gaussnoise}
R\sim \mathcal{N}(\One_D,\lambda  \Sigma)\,,
\end{equation}
where $\One_D$ is the constant $D$-dimensional vector with all entries equal to $1$, $\lambda\geq 0$ is a regularization parameter and $\Sigma \in \RR^{D\times D}$ is a symmetric positive semidefinite covariance matrix. When $\lambda=0$, $R$ is almost surely constant equal to $\One$, and we recover the standard learning without noise injection (\ref{eq:erm}). When $\lambda>0$ and $\Sigma = \Id$ we learn from i.i.d. multiplicative Gaussian noise, a variant of dropout that was proposed by~\citet{srivastava2014dropout} and shown to perform very similarly to dropout. When $\Sigma$ is diagonal but not necessarily constant on the diagonal, the amount of noise can vary among units, but the noise is still independent across units. 

Our focus in this paper is on non-diagonal covariance matrices $\Sigma$, which create correlations between the noise at different units. We call this setting \emph{structured noise injection (SNI)}. In the case of multilayer neural network, it is natural to create correlations within layers, and not between layers, which translates to a block-diagonal structure for $\Sigma$, where each block corresponds to the units of a given layer. We further consider two flavors of SNI.

\subsection{SNI with fixed noise covariance}
The basic flavor of SNI is when we fix the noise covariance $\Sigma$ \emph{a priori} and independently from the data, using for example the structure of the network as prior knowledge. This is a way to input prior knowledge about the problem in the learning algorithm, and has already been proposed as a promising technique in different settings, particularly with binary noise. For example, \citet{tompson_efficient_2014} proposed the SpatialDropout method, where entire feature maps corresponding to adjacent pixels are randomly discarded together instead of individual pixels, corresponding to binary SNI with a block-diagonal covariance matrix with constant blocks equal to $1$ for each set of pixels in a feature map. Similarly, \citet{devries2017improved} applies binary SNI at the input layer of a convolutional network by masking contiguous sections of inputs rather than individual pixels, yielding new state-of-the-art results in image classification. Implementing a SNI strategy with fixed, block-diagonal covariance matrix at the layer level, is only a slight generalization of standard parameter inference with SGD and backpropagation. For example, Algorithm~\ref{algo:SNI} illustrates the forward pass between layers $l-1$ and $l$ with SNI regularization, where $\Sigma^{(l-1)}$ is the block of $\Sigma$ corresponding to layer $l-1$.
\begin{algorithm}[ht]
    \caption{Feed-forward pass with SNI at layer $l$}
    \label{algo:SNI}
    \hspace*{\algorithmicindent} \textbf{Input:} Mini-batch of outputs from the previous $(l-1)$-th layer  $y_1^{(l-1)}, \ldots, y_n^{(l-1)} \in \RR^{d^{(l-1)}}$, regularization parameter $\lambda \in \mathbb{R_+}$, covariance matrix of the noise $\Sigma^{(l-1)}$\\
    \hspace*{\algorithmicindent} \textbf{Output:} The mini-batch of outputs from the $l$-th layer
    \begin{algorithmic}[1] % The number tells where the line numbering should start
	     \For{$i=1$ to $n$}
              \State Sample  $r^{(l-1)}_i \sim \mathcal{N}(\One_{d^{(l-1)}},\lambda  \Sigma^{(l-1)})$ 
              \State  $\widetilde{y}^{(l-1)}_i \gets r^{(l-1)}_i \odot y^{(l-1)}_i$
              \State  $z_{i}^{(l)} \gets W^{(l)}  \widetilde{y}^{(l-1)}_i+ b^{(l)} $
               \State    $y_{i}^{(l)} \gets \sigma(z_{i}^{(l)})$ 
               \EndFor
	\State \textbf{return} $y_1^{(l)}, \ldots, y_n^{(l)}  \in \RR^{d^{(l)}}$
    \end{algorithmic}
\end{algorithm}

\subsection{Adaptive SNI}
Another SNI approach is to define a noise structure with covariance $\Sigma(\Dcal,\theta)$ which may depend on the data and on the model parameters. We refer to this situation as \emph{adaptive structured noise injection (ASNI)}. An example of ASNI approach, where $\Sigma$ depends on the data $\Dcal$ but not on the model parameters $\theta$, was recently proposed by \citet{aydore_using_2018}: they first use the data $\Dcal$ to identify groups of correlated features, and then perform dropout at the group level, reporting promising empirical results. In other words, they impose a block diagonal correlation structure on the noise, where each block corresponds to a group of correlated features, and the correlation of the noise within a group is $1$. While grouping is needed when one wants to perform dropout by group, more general correlation matrices are possible when the noise in Gaussian. An obvious extension of the work of \citet{aydore_using_2018} is to impose over the noise the same covariance structure as observed in the data. When a single-layer linear model is considered, as in \citet{aydore_using_2018}, then $\Sigma$ depends only on the data $\Dcal$, but not on the model parameters $\theta$. In the case of multi-layer networks, the covariance of the units at a given internal layer not only depends on the data distribution, but also on the parameters $\theta$ of the models; as a result, the covariance of the noise in internal layers depends on both $\Dcal$ and $\theta$ when we impose that is equals that of the units in the layer.

To solve (\ref{eq:asni}) with ASNI, we can still follow a SGD approach where at each step a minibatch of examples is randomly chosen, and a realization of the noise on these examples is sampled. One difficulty in ASNI is that the law of the noise depends on the data themselves, and on the parameters of the DNN which evolve during optimization. Similarly to the procedure used in batch normalization~\citep{ioffe2015batch}, we propose to re-estimate the covariance of the noise at each SGD iteration from the mini-batch itself, before sampling the noise on the mini-batch using this structure. Algorithm~\ref{algo:SNI} details the feed-forward pass for one layer, in the case where $\Sigma$ is block diagonal with a block for each layer equal to the covariance of the units in that layer.

We note that in order to sample the noise with a given covariance matrix $\Sigma$, one typically needs to factorize $\Sigma = U U^\top$ by spectral decomposition or Cholesky decomposition \citep{gentle2009computational}, and then get the samples as $U\epsilon$ where $\epsilon \sim \mathcal{N}(\One, \Id)$. This can create significant computational burden, since this must be performed at each SGD iteration, and is one of the reasons why, for example, batch normalization only scales each feature but does not perform whitening \citep{ioffe2015batch}. In the particular case of ASNI where $\Sigma$ is the covariance of the data, we can get an important speed-up since the estimate of $\Sigma$ on a mini-batch is already factorized as $\hat{\Sigma} = \tilde{Y}^\top \tilde{Y}$, where $\tilde{Y}$ is the $n\times d^{(l-1)}$ matrix of mini-batch outputs, centered and divided by $\sqrt{n}$. In other words, the estimation of $\Sigma^{(l-1)}$ in line 1 of Algorithm~\ref{algo:ASNI} can be completely bypassed, and the matrix $U$ in line 4 replaced by $\tilde{Y}$. We note that this is particularly efficient when the mini-batch size $n$ is not larger than the number of units. Finally, a further important speed-up is possible by sampling a single noise vector for each sample in a mini-batch, instead of sampling a different vector for each sample. In other words, we can move lines 3 and 4 of Algorithm~\ref{algo:ASNI} outside of the "for" loop, between lines 1 and 2. This may come at a price of more iteration, since it increases the variance of the stochastic gradient, but we found in our experiments that it did not significantly affect the number of iterations needed and brought a significant overall speed-up.

%\textcolor{blue}{For the generalised structured dropout (SNI) with a general structure matrix $\Sigma$, sampling a random vector from the corresponding dimension-multivariate normal distribution with mean vector $\One$ and covariance $\Sigma$ typically passes by sampling the centered independent Normal and uses the Cholesky decomposition \citep{gentle2009computational}. In the case of (ASNI), however, we already know a decomposition of our matrix that can be used for the sampling. Indeed rotating an independent multivariate Gaussian random vector by the activations matrix of the layer at a pass (estimated using the batch), results in a multivariate Gaussian with covariance $\Sigma$ that is the covariance of the units activations. This results in the algorithm~\ref{algo2:ASNI}\\}
%Note that this is different from applying dropout on rotated activations, since the Hadamard and the matrix product are not interchangeable. Note also that the additional computation for the ASNI algorithm is just a matrix vector  multiplication that can also be easily parallelized. This is also much faster than directly de-correlating layer activations even if it does have a strong decorrelating effect. Note also that unlike the Cholesky decomposition, the sampling procedure is trivially invariant by re-ordering. However, both algorithms eventually converge to the same state.}

\begin{algorithm}[htbp]
    \caption{Feed-forward pass with ASNI at layer $l$}
    \label{algo:ASNI}
    \hspace*{\algorithmicindent} \textbf{Input:} Mini-batch of outputs from the previous layer  $y_1^{(l-1)}, \ldots, y_n^{(l-1)} \in \RR^{d^{(l-1)}}$, regularization parameter $\lambda \in \mathbb{R_+}$ \\
    \hspace*{\algorithmicindent} \textbf{Output:} The mini-batch of outputs from the $l$-th layer
    \begin{algorithmic}[1] % The number tells where the line numbering should start
    \State Estimate $\Sigma^{(l-1)} = U U^\top$ with $U\in\RR^{d^{(l-1)}\times d_U}$ from the batch
%	     \STATE $m^{(l-1)} \gets \frac{1}{n}\sum_{i=1}^n y_i^{(l-1)}$
%	     \STATE $\Sigma^{(l-1)} \gets \frac{1}{n} \sum_{i=1}^{n}(y_{i}^{(l-1)} - m^{(l-1)})(y_{i}^{(l-1)} -   m^{(l-1)} )^\top$ \COMMENT{Empirical covariance of the minibatch}
%	     \STATE Sample  $r^{(l-1)} \sim \mathcal{N}(\One_{d^{(l-1)}},\lambda  \Sigma^{(l-1)})$ 
	     \For{$i=1$ to $n$}           
	     \State Sample  $\epsilon_i \sim \mathcal{N}(\One_{d_u}, \Id)$ 
	     \State $r^{(l-1)}_i \gets \sqrt{\lambda}U \epsilon_i$
              \State  $\widetilde{y}^{(l-1)}_i \gets r^{(l-1)}_i \odot y^{(l-1)}_i$
              \State  $z_{i}^{(l)} \gets W^{(l)}  \widetilde{y}^{(l-1)}_i+ b^{(l)} $
               \State    $y_{i}^{(l)} \gets \sigma(z_{i}^{(l)})$ 
               \EndFor
	\State \textbf{return} $y_1^{(l)}, \ldots, y_n^{(l)}  \in \RR^{d^{(l)}}$
    \end{algorithmic}
\end{algorithm}

%\begin{algorithm}[htbp]
%    \caption{Fast Feed-forward pass with ASNI at layer (l) }
%    \label{algo2:ASNI}
%    \begin{algorithmic}[1] % The number tells where the line numbering should start
%    \REQUIRE Mini-batch of outputs from the previous layer  $y_1^{(l-1)}, \ldots, y_n^{(l-1)} \in \RR^{d^{(l-1)}}$, regularization parameter $�\lambda \in \mathbb{R_+}$
%    \OUTPUT The mini-batch of outputs from the $l$-th layer
%    	     \STATE $m^{(l-1)} \gets \frac{1}{n}\sum_{i=1}^n y_i^{(l-1)}$
%	      \STATE Sample  $r^{(l-1)} \sim  \mathcal{N}(\One_{n},\lambda \Id) $ 
%              \STATE $r^{(l-1)}  \gets \frac{1}{\sqrt{n}} \begin{pmatrix}
%y_{1}^{(l-1)} - m^{(l-1)} 
%\\ 
%\ldots 
%\\ 
%y_{n}^{(l-1)} - m^{(l-1)}
%\end{pmatrix} ^\top   r^{(l-1)} $
%	     \FOR{$i=1$ to $n$}
%              \STATE  $\widetilde{y}^{(l-1)}_i \gets r^{(l-1)} \odot y^{(l-1)}_i$
%              \STATE  $z_{i}^{(l)} \gets W^{(l)}  \widetilde{y}^{(l-1)}_i+ b^{(l)} $
%               \STATE    $y_{i}^{(l)} \gets \sigma(z_{i}^{(l)})$ 
%               \ENDFOR
%	\STATE \textbf{return} $y_1^{(l)}, \ldots, y_n^{(l)}  \in \RR^{d^{(l)}}$
%    \end{algorithmic}
%\end{algorithm}

\section{Regularization effect}
It is well known that learning with noisy data can be related to regularization. For example, adding uncorrelated Gaussian noise to data in ordinary least squares regression is equivalent, in the case of squared error, to ridge regression on the original data \cite{bishop1995training}. The following result clarifies how correlations in the noise impacts this regularization. We consider a setting with no hidden layer and no bias term, i.e., a simple linear model  of the form $f(x)=w^\top x$ where $w\in\RR^d$ is the vector of weights of the model.
\begin{lemma}\label{lem:reg}
Given a training set $(x_1,y_1),\ldots,(x_n,y_n)\in\RR^d\times \RR$ with centred inputs ($\sum_{i=1}^n x_i=0$), and given $R_1,\ldots,R_N$ i.i.d. random variables following $\mathcal{N}(\mu,\lambda \Sigma)$ for some $\lambda\geq 0$ and covariance matrix $\Sigma$, the following holds for any $w\in\RR^d$:
\begin{equation*}
%\begin{split}
\frac{1}{N}  \sum_{i=1}^{N}  \Eb  \br{w^\top(R_i \odot x_i) - y_{i}}^2  
= \frac{1}{N}  \sum_{i=1}^{N}   \br{w^\top x_i - y_{i}}^2 + \lambda w^\top  \br{C \odot \Sigma}  w \,,
%\end{split}
\end{equation*}
where $C = \frac{1}{n}\sum_{i=1}^N x_i x_i^\top$ is the covariance matrix of the data. Furthermore, if $L:\RR^2\rightarrow\RR$ is a general loss function and for any $y\in\RR$, the function $u\in\RR \mapsto \ell_y(u) = L(u,y)$ twice-differentiable, then the following holds:
\begin{equation*}
%\begin{split}
\frac{1}{N}  \sum_{i=1}^{N}  \Eb  L\br{w^\top(R_i \odot x_i) , y_{i}}  
= \frac{1}{N}  \sum_{i=1}^{N}  L \br{w^\top x_i , y_{i}} + \frac{\lambda}{2} w^\top  \br{J(w) \odot \Sigma}  w + o(\lambda)\,,
%\end{split}
\end{equation*}
where
$$
J(w)= \frac{1}{N} \sum_{i=1}^N \ell_{y_i}''\br{w^\top x_i} x_i x_i^\top\,.
$$
\end{lemma}
The proof of these results follows standard arguments~\citep[e.g.][]{wager2013dropout}. They show how the structure in the noise can be interpreted as a particular regularization. Depending on the noise covariance $\Sigma$, we derive several interesting situations:
\begin{itemize}
\item In the case of standard dropout regularization with i.i.d. nodes ($\Sigma=\Id$), we recover known results of \citet{wager2013dropout,baldi2013understanding}.
\item When $\Sigma = \One_d \One_d^\top$, i.e., when the noise is the same for all units, then $\Sigma$ is the neutral multiplication of the Hadamard product. This implies that the regularization boils down to $ \lambda w^\top C w$ in the least squares regression case, and to $ \lambda w^\top J(x) w / 2$ in the more general case. Interestingly, in the least squares regression with centered data, we have the following:
\begin{lemma}\label{lem:zca}
Given a training set $(x_1,y_1),\ldots,(x_n,y_n)\in\RR^d\times \RR$ with centred inputs ($\sum_{i=1}^n x_i=0$), and given $R_1,\ldots,R_N$ i.i.d. random variables following $\mathcal{N}(\mu,\lambda \One_d \One_d^\top)$ for some $\lambda\geq 0$, the following holds for any $w\in\RR^d$:
\begin{equation*}
%\begin{split}
\frac{1}{N}  \sum_{i=1}^{N}  \Eb  \br{w^\top(R_i \odot x_i) - y_{i}}^2  
= \frac{1}{N}  \sum_{i=1}^{N}   \br{\tilde{w}^\top \tilde{x}_i - y_{i}}^2 + \lambda \tilde{w}^\top  \tilde{w} \,,
%\end{split}
\end{equation*}
where $C = \frac{1}{n}\sum_{i=1}^N x_i x_i^\top$ is the covariance matrix of the data, $\tilde{x}_i$ is a whitened version of $x_i$ for $i=1,\ldots,n$, and $\tilde{w}$ is the whitened version of $w$.
\end{lemma}
In other words, SNI on centered data is equivalent in the case of squared error to standard ridge regression on the whitened data. Remember that whitening data is obtained by multiplying each vector by a whitening matrix $Z\in\RR^{d\times d}$ that satisfy $Z^\top Z = C^{-1}$. There exist an infinite number of possible whitening matrix, a standard choice being $Z=C^{-1/2}$ for ZCA whitening. Hence, in Lemma~\ref{lem:zca}, $\tilde{x_i} = Zx_i$ and $\tilde{w} = Zw$, and the proof of Lemma~\ref{lem:zca} results from simple algebric manipulations of the results of Lemma~\ref{lem:reg}. Interestingly, when noise injection is done per layer, then Lemma~\ref{lem:zca} shows that injecting the same noise to all units of a given layer is equivalent to data whitening at the layer input, combined with ridge regularization (a.k.a. weight decay in the neural network terminology). Data whitening is usually associated with high computational cost associated to diagonalization of the covariance matrix, and is replaced in practice by batch normalization which only normalizes the variance of each unit; our analysis suggests that SNI with strong noise correlation within a layer provides a computationally efficient approach to obtain the same result as complete data whitening.
\item Finally, we propose $\Sigma = C$ as another interesting structure for ASNI, which generalizes the idea of \citet{aydore_using_2018} to creating noise correlation on correlated units and in case of non-linear models. In the case of least square regression, this is equivalent by Lemma~\ref{lem:reg} to a regularization by $w^\top C^{\odot 2} w$. While detailed analysis of this choice is complicated by the fact that the eigenstructure of $C^{\odot 2}$ is not easily related to the one of $C$, we show empirically below that it leads to promising results.
\end{itemize}

\section{Effect on learned representation}
While Lemma~\ref{lem:reg} interprets SNI as regularization in the one-layer, linear model case, the same analysis can be done at each layer of a multi-layer network. In that case, though, both the inputs of the layer (i.e., the $x_i$'s in Lemma~\ref{lem:reg}) and the weights $w$ are jointly optimized, since the inputs depend on the parameters of the previous layers. Hence SNI may affect the \emph{representation} learned by a multi-layer network.

Let us take the case of least squares regression as an example, where SNI is equivalent to a regularization by $\lambda w^\top  \br{C \odot \Sigma}  w$ according to Lemma~\ref{lem:reg}. When $\Sigma = \One_d \One_d^\top$, we have seen that SNI is equivalent to ridge regression on the whitened data. Hence, any rotation of the inputs has no impact on the model learned, since both the whitening of the $x_i$'s and the ridge penalty on $w$ are invariant by rotation. As a consequence, SNI in that case does not promote independence of the units in a layer, since any rotation of the units can change the correlation between units without affecting the objective function of SNI regression.

The situation is different for the standard dropout ($\Sigma = \Id$) and ASNI with $\Sigma = C$, as the penalty $w^\top  \br{C \odot \Sigma}  w$ is not invariant by rotation anymore. Interestingly, the following holds:
\begin{lemma}\label{lem:rot}
For $\Sigma = \Id$ or $\Sigma = C$, where $C$ is the covariance matrix of a set of points $x_1,\ldots,x_n \in\RR^d$, it holds that:
\begin{enumerate}
\item $\forall i,j \in [1,d]\,, \br{C \odot \Sigma}_{ij} \geq 0$,
\item $\sum_{i,j=1}^{d}\br{C \odot \Sigma}_{ij}$ is invariant by rotation of the points.
\end{enumerate}
\begin{proof}
For the first point, just notice  that for $\Sigma = \Id$, the entries of $C \odot \Sigma$ are $C_{ii}\geq 0$ on the diagonal, and $0$ elsewhere; for  $\Sigma = C$, the entries of $C \odot \Sigma$ are $C_{ij}^2\geq 0$. For the second point, note that for $\Sigma = \Id$, the sum considered is just the trace of the covariance matrix $C$, which is invariant by rotation (sum of eigenvalues); for $\Sigma=C$, the sum considered it the squared Frobenius norm of $C$, which is also invariant by rotation (sum of the squares of eigenvalues).
\end{proof}
\end{lemma}
As the penalty induced by SNI is
\begin{equation}\label{eq:expand}
\lambda w^\top  \br{C \odot \Sigma}  w = \lambda \sum_{i,j=1}^{d}\br{C \odot \Sigma}_{ij} w_i w_j\,,
\end{equation}
Lemma~\ref{lem:rot} suggests an interplay between the optimization of the inputs (which impact $C\odot\Sigma$) and the optimization of $w$. If we fix $w$ and just optimize over a rotation of the inputs, then the penalty (\ref{eq:expand}) is just a linear function in  $C\odot\Sigma$, which according to Lemma~\ref{lem:rot} stays in a linear polyhedron defined by linear equalities and inequalities, and one might expect the best rotation to push  $C\odot\Sigma$ near the boundary of that polyhedron, where some entries are $0$. Of course a more careful analysis is needed to make this reasoning rigorous (in particular, $w$ should also be rotated, and $C\odot\Sigma$ can not span the whole polyhedron), but it may hints that both dropout and ASNI with $\Sigma=C$ tend to create representations with small values in $C\odot\Sigma$. While this only concerns variance terms for usual dropout, a possible benefit of ASNI with $\Sigma=C$ is that it involves all off-diagonal terms $C_{ij}^2$ as well, suggesting that ASNI may create less correlated representations by penalizing the correlation among units of a given layer. We study this effect more precisely in the experiments below.

\section{Experiments}

\subsection{Simulation}
In order to study the performance of ASNI on a toy model, we use the classical simulation setting proposed by \citet{guyon2007competitive} for the MADELON dataset. In short, we generate 100 samples for training and 10 000 test samples from 2 balanced classes, and train a linear model on the same training set using (1) no dropout, (2)  i.i.d Gaussian dropout with different values of $\lambda$, and (3) ASNI using $\Sigma=C$ with different values of $\lambda$. The MADELON procedure allows to vary total number of features, as well as the number or redundant features. We report in Tables~\ref{tab-auc_cvexp} and~\ref{tab-auc_cvred} the test accuracies of the different models, when we vary the total number of features on the one hand, and when we vary the number of redundant features on the other hand.

\begin{table}[htbp]
\caption{Best average test classification score of a linear model without noise injection, with i.i.d Gaussian dropout, and with ASNI on the MADELON simulation with 10\% useful features and no redundant features: varying the total number of features.}
 \label{tab-auc_cvexp}
 \vskip 0.15in
\begin{center}
\begin{small}
\begin{sc}
%\scalebox{0.6}{
\begin{tabular}{lcccr}
\toprule
\textbf{Redundant }           & \multicolumn{1}{c}{\textbf{No drop.}} & \textbf{i.i.d Gauss.} & \textbf{ASNI } \\ \midrule
$10^2$                       & 66.6 $\pm 2.6$                               & \textbf{68.3}  $\pm 2.0 $               &  68.0 $\pm 2.0 $                              \\ \midrule
$10^3$                        & 68.1  $\pm  2.3$                            & 68.6  $\pm 2.0 $               & \textbf{68.9}  $\pm 2.2$                            \\ \midrule
$10^4$                       & 55.6  $\pm 2.5$                                & 54.9  $\pm 2.8$                & \textbf{56.2}  $\pm 2.6$                                \\
\bottomrule
\end{tabular}
%}
\end{sc}
\end{small}
\end{center}
\vskip -0.1in
\end{table}

\begin{table}[htbp]
\caption{Best average test classification score of a linear model without noise injection, with i.i.d Gaussian dropout, and Structured dropout (ASNI) on the MADELON simulation with 1000 features and 100 useful : varying the number of redundant features \label{tab-auc_cvred}.}
\vskip 0.15in
\begin{center}
\begin{small}
\begin{sc}
%\scalebox{0.6}{
\begin{tabular}{lcccr}
\toprule
\textbf{Features }           & \multicolumn{1}{c}{\textbf{No drop.}} & \textbf{i.i.d Gauss.} & \textbf{ASNI } \\ \midrule
$0$                       & 66.6 $\pm 2.3$                                & \textbf{68.3}  $\pm 2,0 $               &  68.0 $\pm 2.0 $                              \\ \midrule
$100$                          & 65.8  $\pm  1.6$                                & 67.6  $\pm 1.3 $               & \textbf{68.2}  $\pm 1.3$                            \\ \midrule
$800$                        & 68.9  $\pm 1.6$                                 & 69.1  $\pm 1.6$                & \textbf{71.6}  $\pm 1.6$                                \\
\bottomrule
\end{tabular}
%}
\end{sc}
\end{small}
\end{center}
\vskip -0.1in
\end{table}

We notice that in most settings ASNI performs best, particularly when the total number of features grows. This suggests that ASNI acts as an effective regularizer even for linear models. It also significantly stands out in the presence of redundant features. An intuition is that ASNI allows us to use the weights of the redundant features in accordance to the useful features they are created from and minimises prediction disagreement among single weights (since it ties weights in the regularisation). 

\subsection{MNIST}
We now assess the performance of ASNI on image classification, using the classical MNIST benchmark. For simplicity, we train  a network with only 2 dense ReLU-activations hidden layers. We do not expect to obtain state-of-the-art results as we do not perform any data augmentation or other regularisation. The goal of this set of experiments is mainly to study the difference of performance and the effect of ASNI on a hidden layer activations compared with independent noise injection. The number of the second hidden layer units $d^{(2)}$ is fixed to the number of classes ($10$ here), and we vary the number of units in the first hidden layer $d^{(1)}$.

\begin{table}[htbp]
\caption{Best average classification score on the MNIST dataset of a 2 hidden layer without noise injection, with i.i.d noise injection (Gaussian and Bernoulli dropout), and with ASNI, when we vary the number of units in the first hidden layer. \label{tab-class_mnist}}
\vskip 0.15in
\begin{center}
\begin{small}
\begin{sc}
\scalebox{1}{
\begin{tabular}{lcccr}
\toprule
\textbf{$d^{(1)}$ }           & \multicolumn{1}{c}{\textbf{No drop.}} & \textbf{iid Gauss.}  & \textbf{iid Bern.} & \textbf{ASNI} \\ \midrule
32	   		& 93.7 $\pm$ 0.2                              & 94.2 $\pm$ 0.8    & 94.4 $\pm$ 0.9               & \textbf{95.8} $\pm$ 0.4                                 \\ \midrule
64                        & 95.8  $\pm $ 0.6                                & 95.4   $\pm$  0.7    & 95.9   $\pm$  0.6               &  \textbf{96.6}  $ \pm$  0.7                                 \\ \midrule
256                        & 96.1  $\pm$  0.6                                & 97.0  $\pm$  0.7      & 97.4  $\pm$  0.7               &  \textbf{97.8}  $\pm$  0.7                                  \\ \midrule
512                        & 96.5  $\pm$  0.1                                 & 97.5 $\pm$  0.1     & 97.6 $\pm$  0.1               &  \textbf{98.1} $\pm$  0.1                                 \\ \midrule
1,024		  & 96.2  $\pm$  0.1                            	      & 97.6 $\pm$  0.1     & 97.6 $\pm$  0.2               &  \textbf{98.1} $\pm$ 0.3                                \\ 
\bottomrule
\end{tabular}
}
\end{sc}
\end{small}
\end{center}
\vskip -0.1in
\end{table}

We summarize in Table~\ref{tab-class_mnist} the test accuracy defined as the proportion of well classified examples from the test set, after training the 2 hidden layer network with varying number of units. Figure~\ref{fig:mnist_acc_varying} shows the evolution of this test error during the training process of the model with 256 units in the hidden layer, as a function of the number of SGD iterations. 

We see from Table~\ref{tab-class_mnist} that all methods involving noise injection tend to outperform the baseline approach without no regularization, which confirms the benefits of noise injection for performance. Second, we notice that among the three methods that perform noise injection, ASNI constantly outperforms both Gaussian and Bernoulli i.i.d dropout for small and large number of units. The 2 hidden layers, that seems to overfit even for a small number of units in both hidden layers, has however a better accuracy with larger number of units, which indicates that there is still information to be gained from the data. The network with 64 units however, with ASNI regularization, seems to capture more information than the network without dropout with 1024 units, which can be largely explained by the quality of representation learnt by structured dropout, as we will show below. 
Figure \ref{fig:mnist_acc_varying} also shows that ASNI leads to faster convergence, an effect observed as well in batch normalisation \citep{ioffe2015batch}.

\begin{figure}[ht]
\vskip 0.2in
\begin{center}
\centerline{
  \includegraphics[scale=0.37]{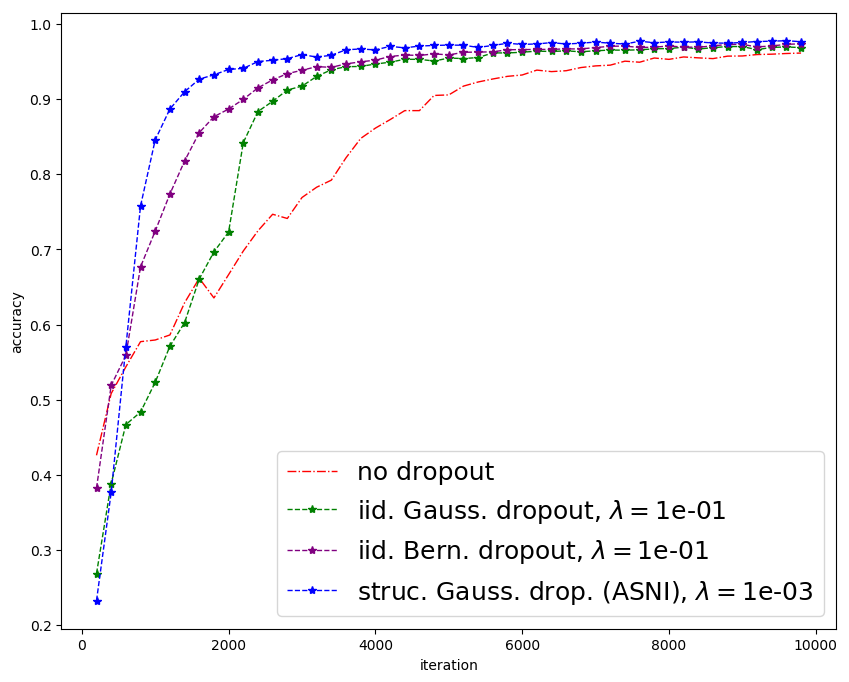}}
\caption{Test classification during training for a 2-hidden layers MLP trained without noise injection, with i.i.d. noise injection or with ASNI (with the best regularisation hyper-parameter), with 256 units in the hidden layer. \label{fig:mnist_acc_varying}}
\end{center}
\vskip -0.2in
\end{figure}

To see the effects of ASNI on units co-adaptation we measure the total correlation of the units' activations at a layer $l$  as the Frobenius norm of the activations correlations matrix $T$ defined as: 
$$ T_{ij} \frac{\Sigma_{i,j}}{\sqrt{\Sigma_{i,i} \Sigma_{j,j}}} \,,$$
where $\Sigma$ is the covariance of the unit activations introduced in methods description. The evolution of this quantity is shown in Figure~\ref{fig:mnist_acc_varying}, for the network with 1,024 units in the hidden layer. 
\begin{figure}[ht]
\vskip 0.2in
\begin{center}
\centerline{
  \includegraphics[scale=0.37]{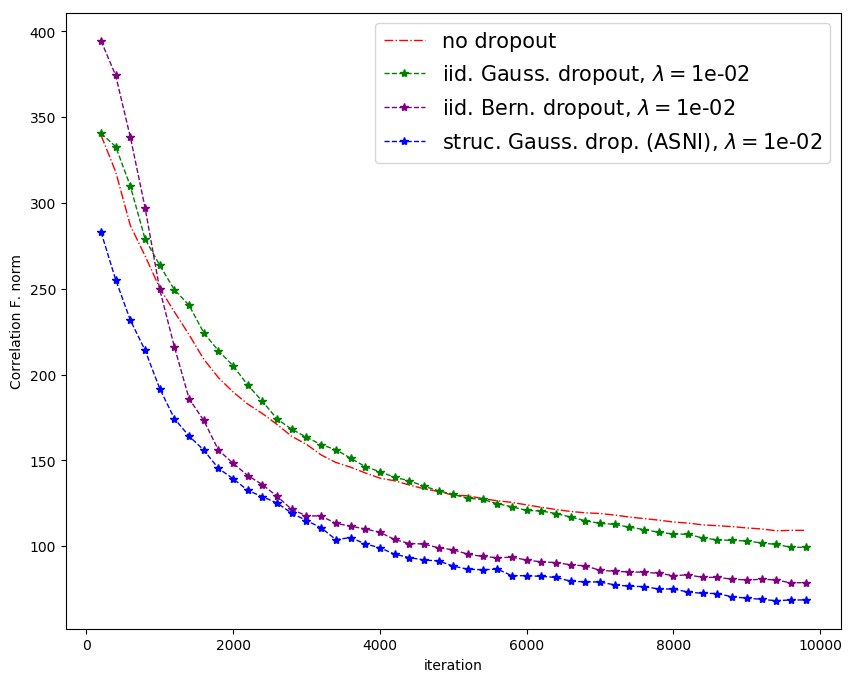}
} 
\caption{Correlation matrix norm of the first hidden layer activations during training for a 2-hidden layers MLP trained without noise injection, with i.i.d. noise injection or with ASNI (with the best regularisation hyper-parameter), with 1,024 units in the first hidden layer. \label{fig:mnist_acc_varying}}
\end{center}
\vskip -0.2in
\end{figure}
We see that for all methods, correlations among units tends to decrease during optimization, which confirms that better performance is obtained when units are less redundant. We also see that adding noise has a dramatic effect on the decrease of correlation, as all three methods regularized by noise injection see their units' correlations decrease much faster and much lower than the unregularized baseline. Gaussian and Bernoulli i.i.d. noise injection lead to a very similar curve, confirming that both methods behave very similarly. Finally, we observe that ASNI lead s to faster decrease of the correlation matrix norm, and that it reaches after $10^4$ iterations a lower value than all other methods. This empirically confirms the active role played by non-i.i.d. noise injection, in particular ASNI, in promoting non-redundant representations.

\begin{figure}[!htb]
\centering{
  \includegraphics[scale=0.3]{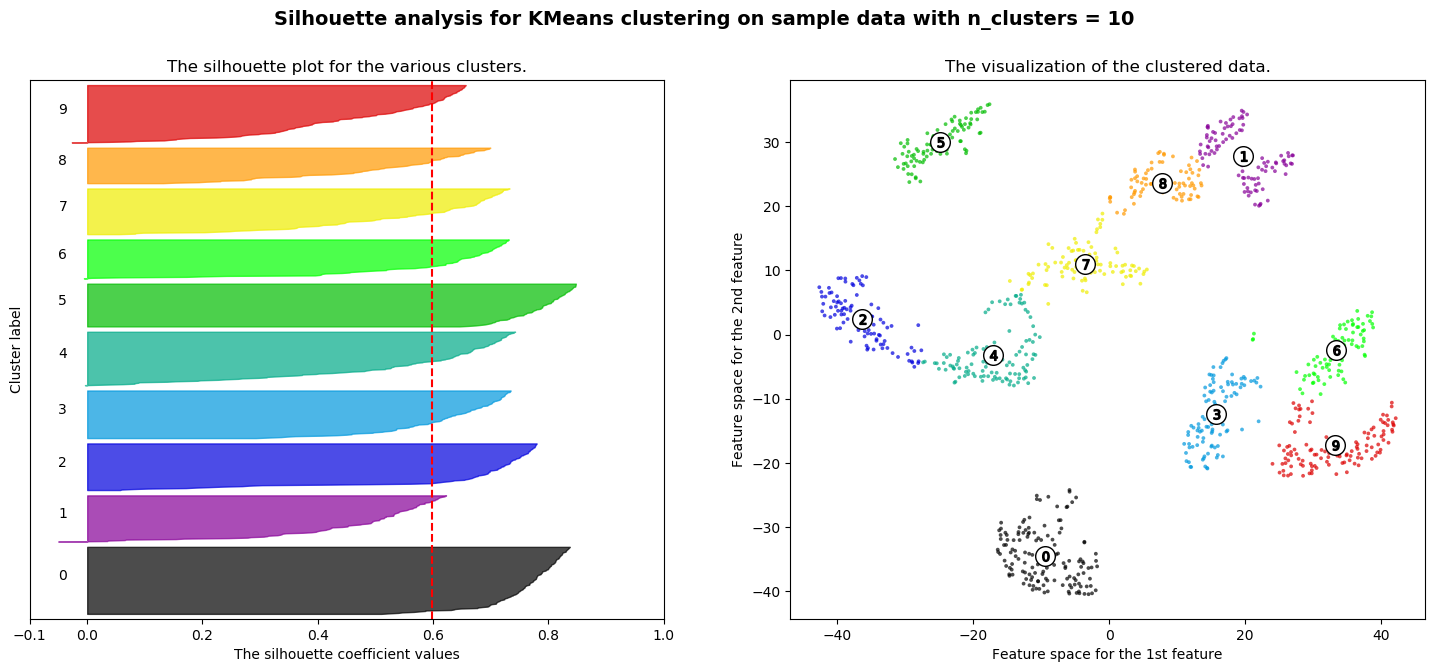}
  \includegraphics[scale=0.3]{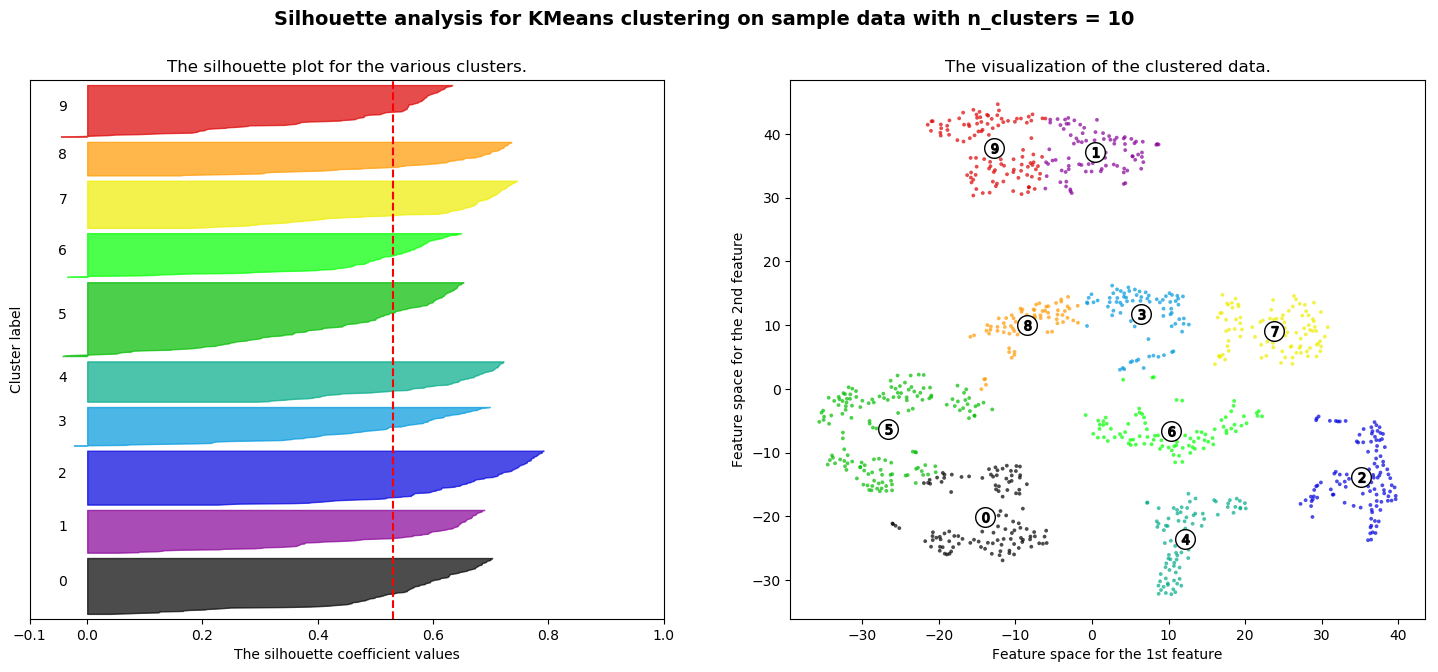}
  \includegraphics[scale=0.3]{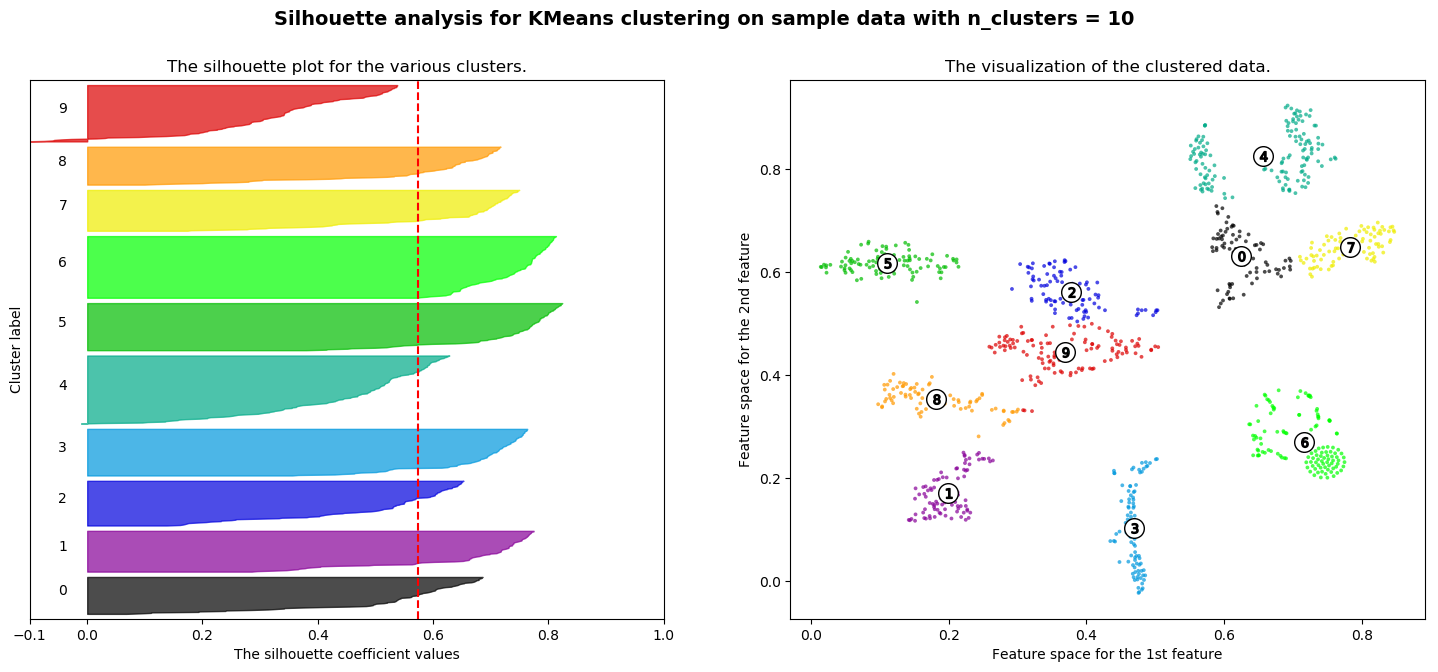}
  \includegraphics[scale=0.3]{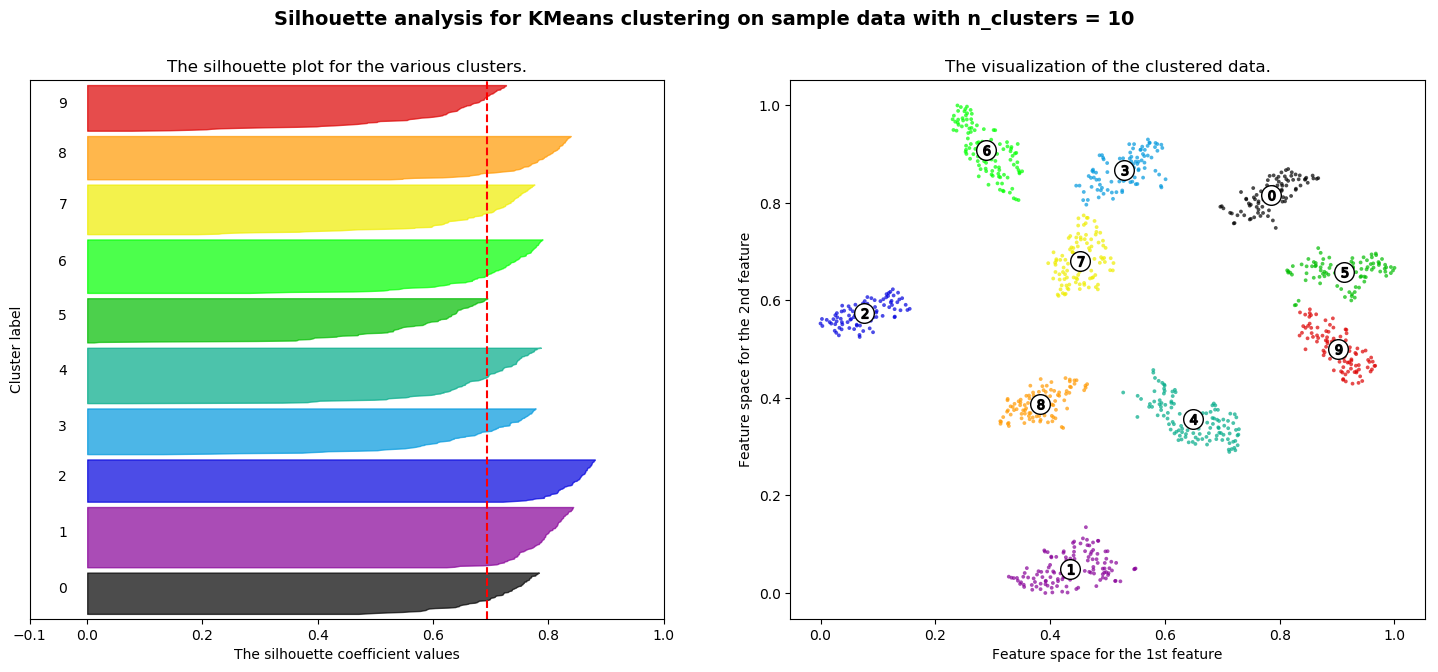}
}
  \caption{Silhouette plots (left) for the t-SNE embeddings (right) of the first hidden layer activations on 1,000 MNIST test images (2 hidden layers MLP with 32 units on the first layer). We compare, from the top to the bottom row, a network trained without noise injection, with i.i.d. Gaussian dropout, with i.i.d Bernoulli dropout, and with ASNI. The points are colored and numbered according to the class of the images.\label{fig:tsne_sil_32}}
\end{figure}

To further study the quality of the representations learned by different methods, we visualize the vectors of hidden layer activations on the test set using t-SNE in order to assess how well the different classes are separated. Figure~\ref{fig:tsne_sil_32} (right panels) shows the 2-component t-SNE embeddings of the second hidden layer activations (32 in this case) applied on a sample of 1,000 test samples, trained respectively without noise injection, with i.i.d. Gaussian or Bernoulli dropout, and with ASNI.
%\begin{figure}[ht]
%\vskip 0.2in
%\begin{center}
%%\centerline{
%  \includegraphics[scale=0.3]{../asni_report/mnist_32_tsne_no.png}
%  \includegraphics[scale=0.3]{../asni_report/mnist_32_tsne_iid.png}
%  \includegraphics[scale=0.3]{../asni_report/mnist_32_tsne_struc.png}
%%} 
%\caption{t-SNE visualization of the second hidden layer activations for 1,000 MNIST test images, for a 2 hidden layers MLP with 32 units on the 1st layer. We compare a network trained without noise injection (upper left), with i.i.d Bernoulli dropout (upper right), and with ASNI (bottom). The points are colored according to the class of the images.\label{fig:tsne_mnist}}
%\end{center}
%\vskip -0.2in
%\end{figure}
Visually, we see that the class are better separated in the representation learned by ASNI than by the other methods. To quantify this visual impression, we measure the quality of the representation by computing the Silhouette coefficient of each t-SNE embedding \citep{rousseeuw1987silhouettes}. A larger Silhouette value indicates that the representation is better at recovering the known classes of images \citep{chen2002evaluation}. We report in Table~\ref{tab-sne_mnist} the mean silhouette coefficients over all test samples for 10 clusters with respectively 32, 256 and 1,024 units in the first hidden layer. These results confirm what we qualitatively observ in Figure~\ref{fig:tsne_sil_32}, namely, that noise injection improves the quality of the representations compared to the non-regularized version, and more importantly that ASNI clearly outperforms i.i.d. noise injection in all settings.
%An example of the Silhouette analysis on MNIST with the 2 hidden layers MLP with 32 units on the 1st layer is visualized in Figure \ref{fig:tsne_sil_32}.
\begin{table}[htbp]
\caption{Average Silhouette coefficient scores of the last hidden layer t-SNE embeddings on MNIST test dataset, without noise injection, with i.i.d Gaussian and Bernoulli dropout, and structured dropout (ASNI). \label{tab-sne_mnist}.}
\vskip 0.15in
\begin{center}
\begin{small}
\begin{sc}
%\scalebox{0.6}{
\begin{tabular}{lcccr}
\toprule
\textbf{$d^{(1)}$ }           & \multicolumn{1}{c}{\textbf{No drop.}} & \textbf{iid Gauss.}  & \textbf{iid Bern.} & \textbf{ASNI} \\ \midrule
32	   		&  0.60                             & 0.57   & 0.53              & \textbf{0.69}                                 \\ \midrule
256                        & 0.58                            & 0.63    &0.63              &  \textbf{0.72}                              \\ \midrule
1024		  &       0.58                 	      &0.73    &     0.74          &  \textbf{0.80}                                \\ 
\bottomrule
\end{tabular}
%}
\end{sc}
\end{small}
\end{center}
\vskip -0.1in
\end{table}

\begin{figure}[htbp]
\centering{
  \includegraphics[scale=0.35]{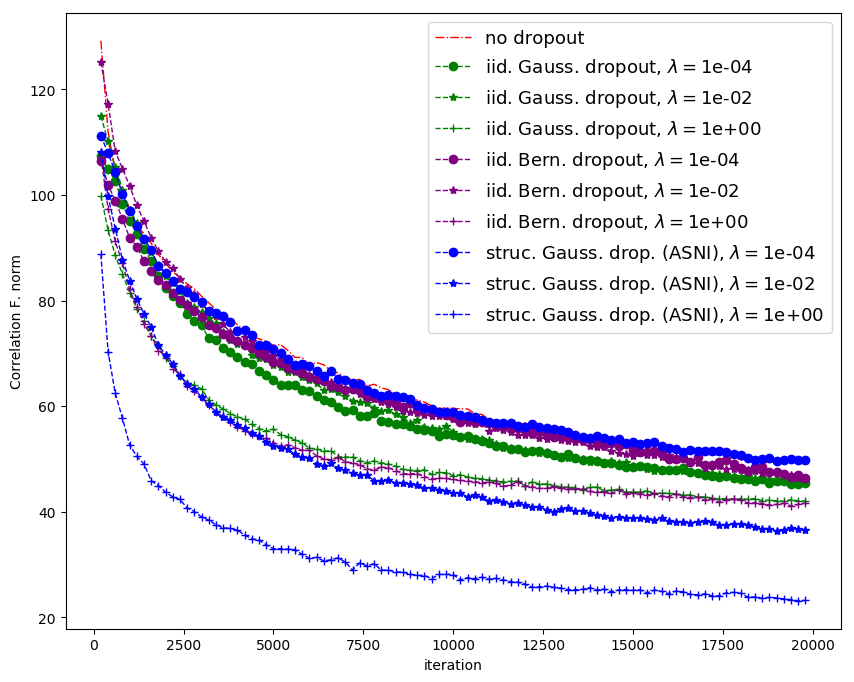}
    \includegraphics[scale=0.35]{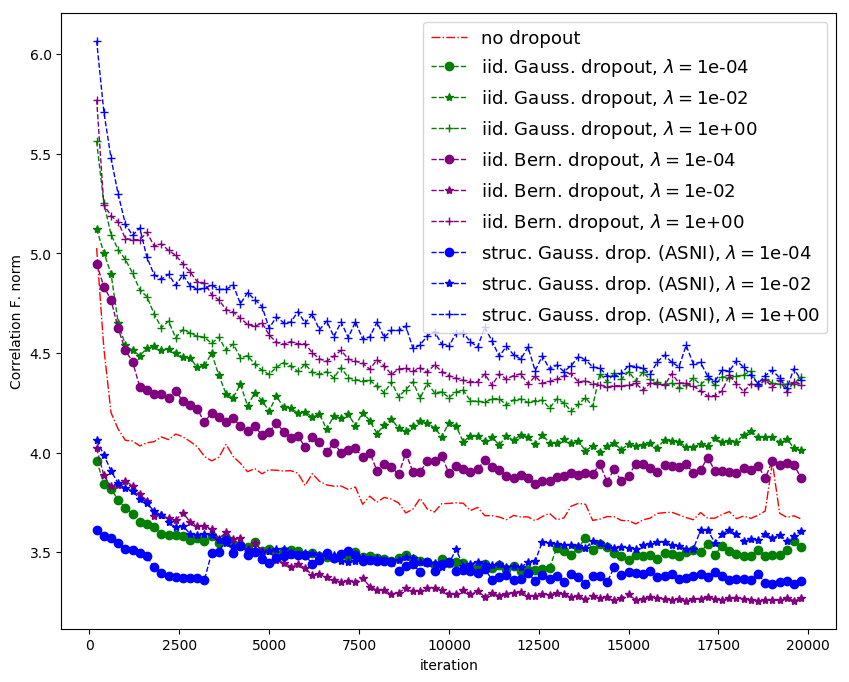}
  \includegraphics[scale=0.35]{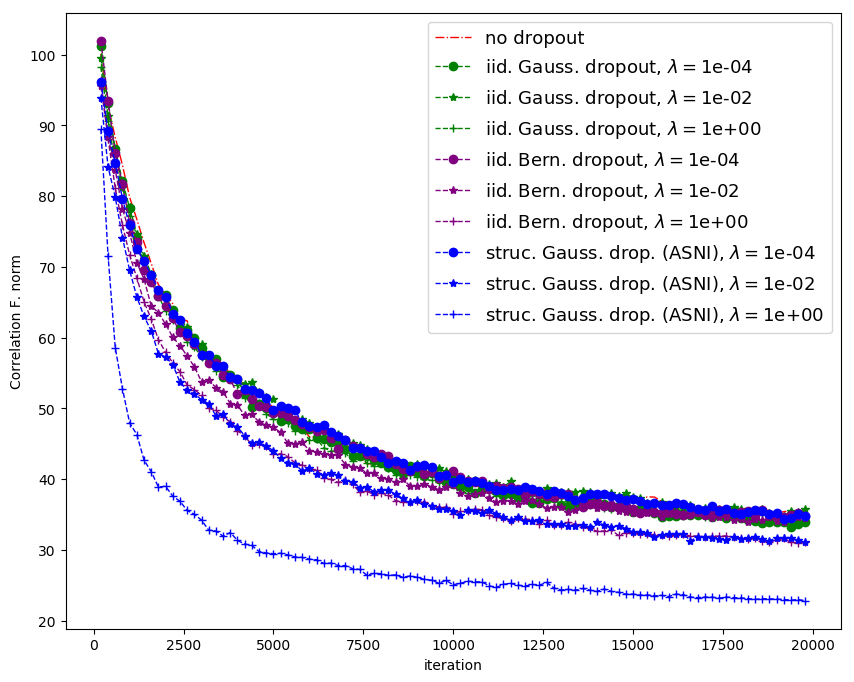}
  \includegraphics[scale=0.35]{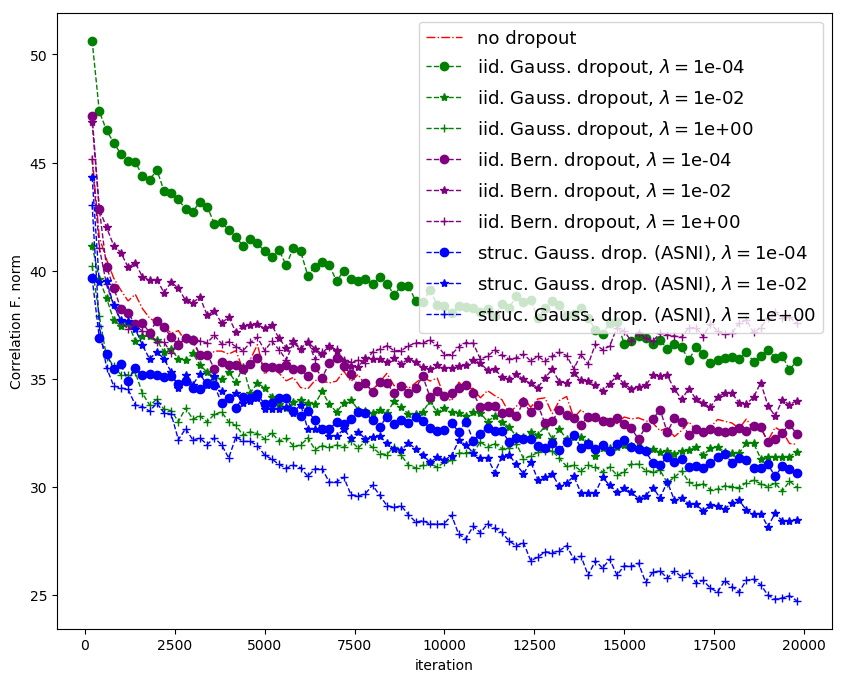}
   \includegraphics[scale=0.35]{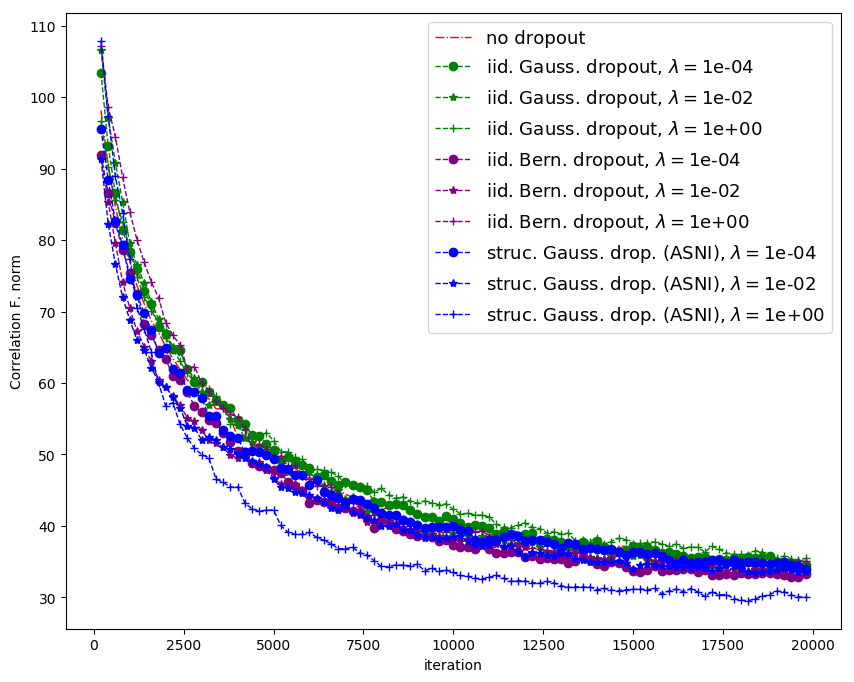}
  \includegraphics[scale=0.35]{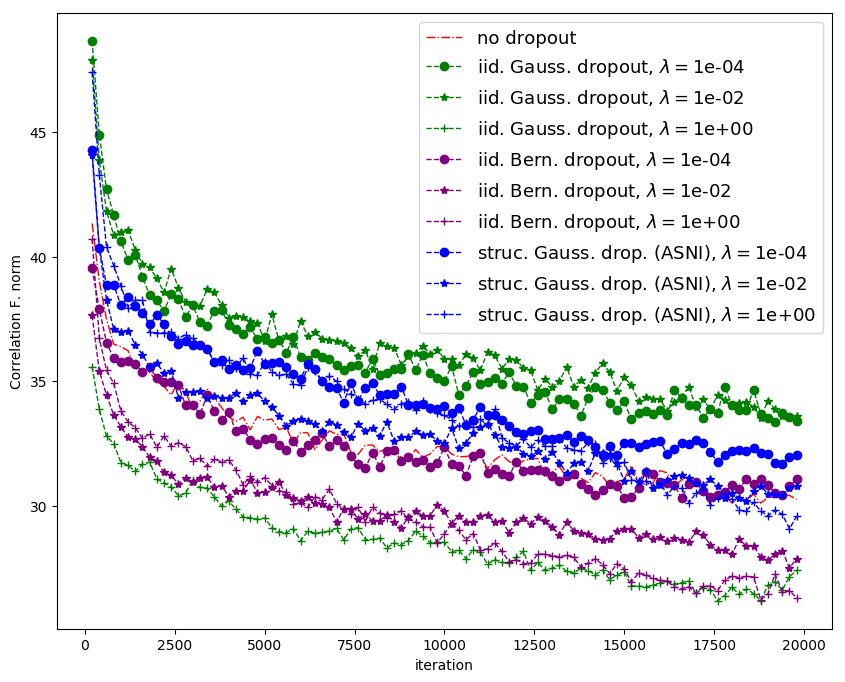}
}
\caption{Correlation matrix norm of the first (left figures) and second (right figures) hidden layer activations during training for a 2-hidden layers MLP with no noise injection, with iid noise injection and ASNI, applied on the first hidden layer only (above), on the the second hidden layer only (middle) or on the input layer only (below).  \label{fig:plot_all_drop}}
\end{figure}

In order to further investigate which layers are decorrelated by ASNI, we train the same 2 hidden layers architecture on MNIST,  but we apply independent noise injection or ASNI only on a single layer for each experiment (input, first hidden or second hidden layer). The evolution of the first and second hidden layer's correlations during training (represented again by the Frobenius norm of the activations correlation matrix at iteration $t$) for each experiment is shown in Figure \ref{fig:plot_all_drop}.
We see that i.i.d. Bernoulli and Gaussian dropout do not necessarily reduce the correlations between units, and thus do not always prevent co-adaptations in terms of activations correlations. ASNI, on the other hand, forces units to be more independent when it is applied on that layer, but does not reduce cross-correlations completely to 0 since the norm of the correlation matrix continues to decrease during the training. In this sense, ASNI is different from the whitening techniques mentioned in the introduction in that it does not explicitly change the input and  does not force units to be independent such as batch normalisation and its decorrelated variants, but rather encourages units through the structure of dropout to be more independent. Interestingly, Figure \ref{fig:plot_all_drop} also shows that in the case of dense multilayer networks, applying ASNI on one layer does decorrelate the activations of that layer but not of the next layers. However, applying ASNI on the second hidden layer decorrelates both its activations and the activations of the first hidden layer.    

\begin{figure}[htbp]
\vskip 0.2in
\begin{center}
%\centerline{}
  \includegraphics[scale=0.8]{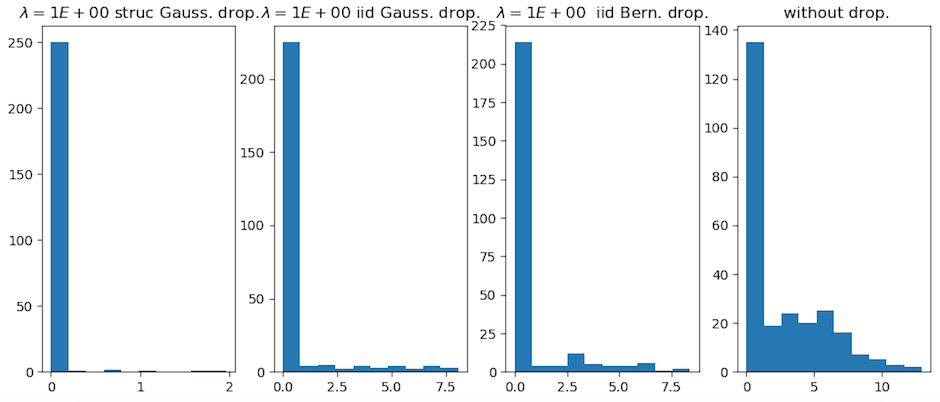}
  \includegraphics[scale=0.8]{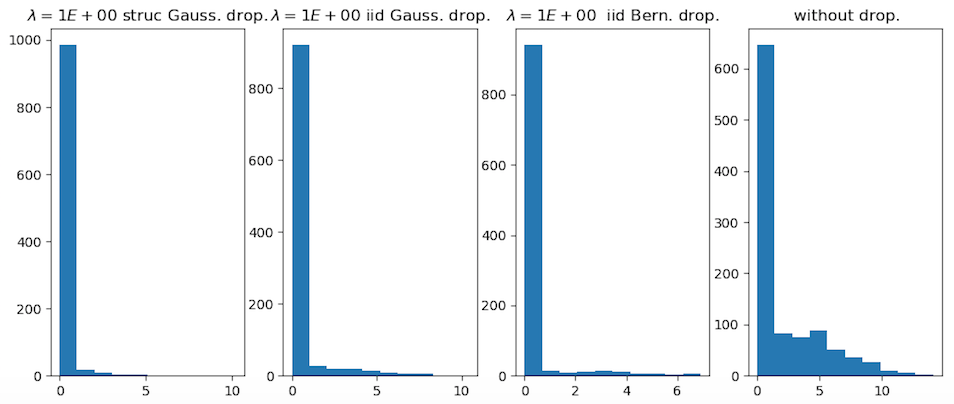} 
\caption{First layer activations after training our 2 hidden layers network on MNIST, without dropout, with i.i.d. gaussian dropout, i.i.d. Bernoulli dropout or structured dropout (ASNI). With 256 units (above) and 1024 units (below).  
\label{fig:sparse_mnist}}
\end{center}
\vskip -0.2in
\end{figure}

Finally, we assess whether ASNI regularization has an effect on the sparsity of activations. Figure \ref{fig:sparse_mnist} shows the histogram of the activation values after training the same 2 hidden layers network without dropout, with i.i.d. gaussian or Bernoulli dropout, or with ASNI. It confirms the findings of \citet{srivastava2014dropout} that dropout may lead to sparser representations. However, we can see that ASNI provides a sparser activations distribution than dropout, while improving on accuracy as previously shown in Table \ref{tab-class_mnist}. We also notice that Bernoulli dropout and its gaussian variant result in a similar level of sparsity, in this experiment at least, which leads to think that this effect is independent from the sparsity of the multiplicative noise itself.

\subsection{CIFAR10 and CIFAR100}

Finally, we compare ASNI to i.i.d. noise injection on a more realistic setting, namely, a LeNet convolutional network architecture with 4 convolutional layers followed by 2 dense layers tested on the CIFAR10 and CIFAR100 datasets.
We again compare the different noise injection schemes applied on the 2 dense hidden dense layers, without data augmentation or additional regularization. 

Table~\ref{tab-class_cifar10} summarizes the test accuracy reached by the different training procedures. We again observe that all noise injection methods outperform the baseline, that Gaussian and Bernoulli i.i.d. dropout behave very similarly, and that ASNI has the best performance for these datasets. We also notice that ASNI has less variance in performance compared to all other methods, which might be explained by the faster convergence observed already in MNIST experiments.      

\begin{table}[htbp]
\caption{Best average test classification score of LeNet on CIFAR10 without noise injection, with i.i.d. noise injection (Gaussian and Bernoulli dropout), and with ASNI on CIFAR10 and CIFAR100 benchmarks.  \label{tab-class_cifar10}.}
\vskip 0.15in
\begin{center}
\begin{small}
\begin{sc}
\scalebox{1}{
\begin{tabular}{lcccr}
\toprule
\textbf{Data}          & \multicolumn{1}{c}{\textbf{No drop.}} & \textbf{iid Gauss.}  & \textbf{iid Bern.} & \textbf{ASNI} \\ \midrule
 CIFAR10		& 66.5  $\pm$  0.1                                 & 67.9 $\pm$  0.3       & 67.7   $\pm$  0.4             & \textbf{68.3  $\pm$  0.2  }                               \\ \midrule
  CIFAR100		& 32.9  $\pm$  0.2                         &33.8  $\pm$  0.5     & 33.8  $\pm$  0.5               & \textbf{34.4  $\pm$  0.3  }         \\  
\bottomrule
\end{tabular}
}
\end{sc}
\end{small}
\end{center}
\vskip -0.1in
\end{table}

%\begin{table}[htbp]
%\caption{Best average test classification score of LeNet on CIFAR100 without noise injection, with i.i.d noise injection, and structured dropout (ASNI)..  \label{tab-class_cifar100}.}
%\vskip 0.15in
%\begin{center}
%\begin{small}
%\begin{sc}
%\scalebox{0.6}{
%\begin{tabular}{lcccr}
%\toprule
%\textbf{CIFAR100}        & \multicolumn{1}{c}{\textbf{No drop.}} & \textbf{iid Gauss. drop.}  & \textbf{iid Bern. drop.} & \textbf{ASNI} \\ \midrule
%   Test accuracy		& 32,86  $\pm$  0.2   \%                          &33,78  $\pm$  0.5 \%     & 33,79  $\pm$  0.5  \%              & \textbf{34,44  $\pm$  0.3  \% }         \\ 
%\bottomrule
%\end{tabular}}
%\end{sc}
%\end{small}
%\end{center}
%\vskip -0.1in
%\end{table}

\begin{figure}[ht]
\vskip 0.2in
\begin{center}
%\centerline{
  \includegraphics[scale=0.35]{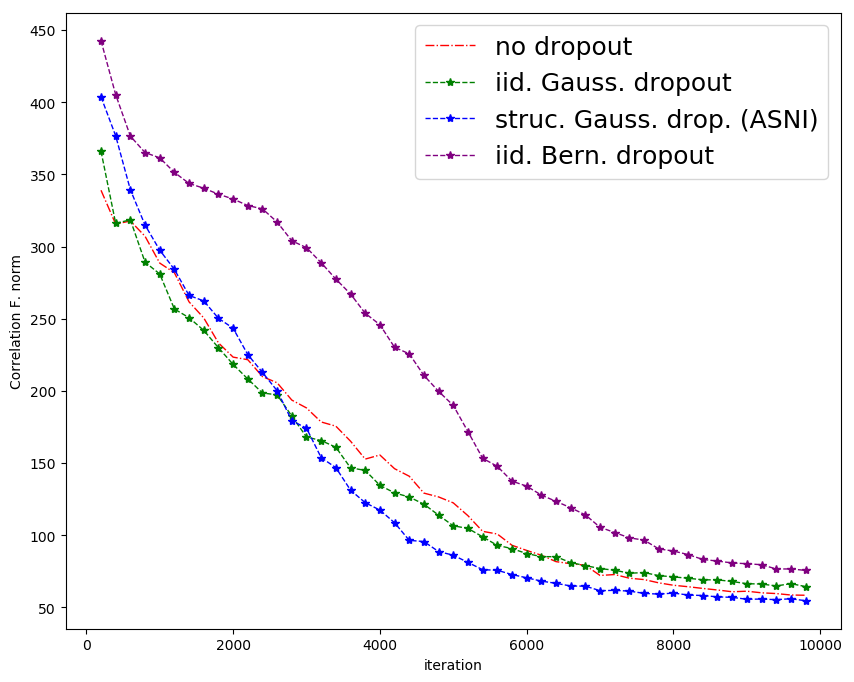}
  \includegraphics[scale=0.35]{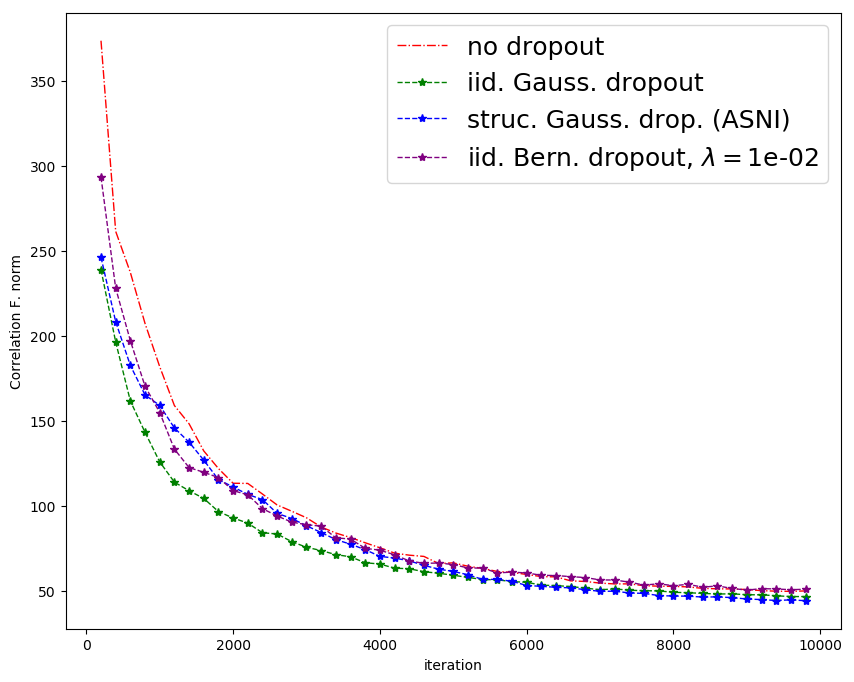}
%} 
\caption{Correlation matrix norm of the first dense hidden layer activations with LeNet, with either no noise injection, iid noise injection or ASNI, during training on CIFAR10 (left) and CIFAR100 (right).  \label{fig:cifar10_cor}}
\end{center}
\vskip -0.2in
\end{figure}

As for the MNIST experiments, we also measure the amount of correlations between unit activations, evaluated by the Frobenius norm of the correlation matrix, and show how it evolves over training for the different methods in Figure~\ref{fig:cifar10_cor}. We notice that standard Bernoulli dropout has a weaker effect on reducing correlations than other methods on CIFAR10, but that overall all methods significantly reduce correlation during training. After convergence, ASNI keeps a small advantage on both datasets in terms of correlation level reached. As shown in Table~\ref{tab-sne_cifar}, the representation learned by ASNI has also a larger Silhouette than other methods on the test sets.
%One observation here is that classical dropout and its Gaussian variant do not necessarily reduce correlations, an observation mentioned already in \citet{helmbold2017surprising}. ASNI, however, seems to always have its disentengeing property in convolutional networks for each layer it is applied on even for a small regularization parameter, while improving the representations power as shown by table \ref{tab-sne_cifar}.

\begin{table}[h]
\caption{Average Silhouette coefficient of LeNet's last hidden layer t-SNE embeddings on CIFAR test datasets, without noise injection, with i.i.d Gaussian and Bernoulli dropout, and with ASNI. \label{tab-sne_cifar}.}
\vskip 0.15in
\begin{center}
\begin{small}
\begin{sc}
\scalebox{1}{
\begin{tabular}{lcccr}
\toprule
\textbf{Data }           & \multicolumn{1}{c}{\textbf{No drop.}} & \textbf{iid Gauss.}  & \textbf{iid Bern.} & \textbf{ASNI} \\ \midrule
CIFAR10	   		&  0.38                             & 0.43   & 0.42              & \textbf{0.48}                                 \\ \midrule
CIFAR100                        & 0.35                            & 0.37    &0.36              &  \textbf{0.38}                              \\ \bottomrule
\end{tabular}
}
\end{sc}
\end{small}
\end{center}
\vskip -0.1in
\end{table}

\section{Conclusion}
We proposed new regularization schemes that generalize dropout, by creating correlations between the noise components. We focused particularly on ASNI, an adaptive approach that replicates the structure of the data correlation in the noise correlation. We showed both theoretical and empirical results suggesting that ASNI improves the representation and performance of shallow and deep neural network, while maintaining computation efficiency. The ASNI framework opens new research directions. First, one way consider different ways to create the noise correlation structure, using for example the structure of the network, or may even think about \emph{learning} it. Second, while Gaussien noise is convenient to impose a particular correlation structure, discrete noises such as binary noise can be computationally advantageous; sampling binary random variables with a given covariance matrix is however not an easy task~\citep{leisch1998generation,preisser2014comparison}, and progress in that direction may be directly useful for DNN regularization.

%We studied ASNI schemes using Gaussian multiplicative noise~(\ref{eq:gaussnoise}), mostly because sampling such noise is simple. It would be interesting to consider other structured noise, for example with Bernouilli or Poisson marginal distributions; however, generating for example binary variables with a given mean and correlation matrix is not a straightforward task~\citep{leisch1998generation}, and we do not even know for which covariance matrices $\Sigma$ such a distribution exists~\citep{preisser2014comparison}. We think that sampling noise based on Monte Carlo Markov Chaines (MCMC) could be an interesting direction to pursue, as it would nicely fit the SGD schemes we use to learn with noise.
%\textcolor{blue}{Finally, we also think that Structured Noise Injection opens the way for experimenting with other structures that might be of interest in other applications. It seems indeed that creating a duality between noise structure and model information can be very interesting. An example that can be explored in future work is when we want a certain structure for hidden layer units that is either known or imposed by the problem context and invariances such as graph-related learning tasks. }

\section{Availability}
All code concerning the real data experiments is available at \url{https://github.com/BeyremKh/ASNI}\\

\clearpage
\beginsupplement

\clearpage

\bibliographystyle{plainnat}
\bibliography{asni}

\begin{thebibliography}{35}
\providecommand{\natexlab}[1]{#1}
\providecommand{\url}[1]{\texttt{#1}}
\expandafter\ifx\csname urlstyle\endcsname\relax
  \providecommand{\doi}[1]{doi: #1}\else
  \providecommand{\doi}{doi: \begingroup \urlstyle{rm}\Url}\fi

\bibitem[Aydore et~al.(2018)Aydore, Thirion, Grisel, and
  Varoquaux]{aydore_using_2018}
Sergul Aydore, Bertrand Thirion, Olivier Grisel, and Gael Varoquaux.
\newblock Using feature grouping as a stochastic regularizer for
  high-dimensional noisy data.
\newblock Technical Report 1807.11718, arXiv, 2018.
\newblock URL \url{http://arxiv.org/abs/1807.11718}.

\bibitem[Ba and Frey(2013)]{ba2013adaptive}
Jimmy Ba and Brendan Frey.
\newblock Adaptive dropout for training deep neural networks.
\newblock In C.J.C. Burges, L.~Bottou, M.~Welling, Z.~Ghahramani, and K.Q.
  Weinberger, editors, \emph{Adv. Neural. Inform. Process Syst.}, volume~26,
  pages 3084--3092. Curran Associates, Inc., 2013.

\bibitem[Baldi and Sadowski(2013)]{baldi2013understanding}
Pierre Baldi and Peter~J Sadowski.
\newblock Understanding dropout.
\newblock In C.J.C. Burges, L.~Bottou, M.~Welling, Z.~Ghahramani, and K.Q.
  Weinberger, editors, \emph{Adv. Neural. Inform. Process Syst.}, volume~26,
  pages 2814--2822. Curran Associates, Inc., 2013.

\bibitem[Barlow et~al.(1959)]{barlow1959possible}
Horace~B Barlow et~al.
\newblock Possible principles underlying the transformations of sensory
  messages.
\newblock \emph{Sensory Communication, Contributions: Contributions}, 217,
  1959.

\bibitem[Bengio et~al.(2013)Bengio, Courville, and
  Vincent]{bengio_representation_2012}
Yoshua Bengio, Aaron Courville, and Pascal Vincent.
\newblock Representation learning: a review and new perspectives.
\newblock Technical Report~8, 2013.
\newblock URL \url{http://dx.doi.org/10.1109/TPAMI.2013.50}.

\bibitem[Bishop(1995)]{bishop1995training}
Chris~M Bishop.
\newblock Training with noise is equivalent to tikhonov regularization.
\newblock \emph{Neural computation}, 7\penalty0 (1):\penalty0 108--116, 1995.

\bibitem[Chen et~al.(2002)Chen, Jaradat, Banerjee, Tanaka, Ko, and
  Zhang]{chen2002evaluation}
Gengxin Chen, Saied~A Jaradat, Nila Banerjee, Tetsuya~S Tanaka, Minoru~SH Ko,
  and Michael~Q Zhang.
\newblock Evaluation and comparison of clustering algorithms in analyzing es
  cell gene expression data.
\newblock \emph{Statistica Sinica}, pages 241--262, 2002.

\bibitem[Cogswell et~al.(2015)Cogswell, Ahmed, Girshick, Zitnick, and
  Batra]{cogswell2015reducing}
Michael Cogswell, Faruk Ahmed, Ross Girshick, Larry Zitnick, and Dhruv Batra.
\newblock Reducing overfitting in deep networks by decorrelating
  representations.
\newblock Technical report, 2015.

\bibitem[Desjardins et~al.(2015)Desjardins, Simonyan, Pascanu,
  et~al.]{desjardins2015natural}
Guillaume Desjardins, Karen Simonyan, Razvan Pascanu, et~al.
\newblock Natural neural networks.
\newblock In \emph{Advances in Neural Information Processing Systems}, pages
  2071--2079, 2015.

\bibitem[DeVries and Taylor(2017)]{devries2017improved}
Terrance DeVries and Graham~W Taylor.
\newblock Improved regularization of convolutional neural networks with cutout.
\newblock Technical report, 2017.

\bibitem[Dietterich(2000)]{dietterich2000ensemble}
Thomas~G Dietterich.
\newblock Ensemble methods in machine learning.
\newblock In \emph{International workshop on multiple classifier systems},
  pages 1--15. Springer, 2000.

\bibitem[Gal and Ghahramani(2016)]{gal2016theoretically}
Yarin Gal and Zoubin Ghahramani.
\newblock A theoretically grounded application of dropout in recurrent neural
  networks.
\newblock In \emph{Advances in neural information processing systems}, pages
  1019--1027, 2016.

\bibitem[Gentle(2009)]{gentle2009computational}
James~E Gentle.
\newblock \emph{Computational statistics}, volume 308.
\newblock Springer, 2009.

\bibitem[Guyon et~al.(2007)Guyon, Li, Mader, Pletscher, Schneider, and
  Uhr]{guyon2007competitive}
Isabelle Guyon, Jiwen Li, Theodor Mader, Patrick~A Pletscher, Georg Schneider,
  and Markus Uhr.
\newblock Competitive baseline methods set new standards for the nips 2003
  feature selection benchmark.
\newblock \emph{Pattern recognition letters}, 28\penalty0 (12):\penalty0
  1438--1444, 2007.

\bibitem[Hassibi and Stork(1993)]{hassibi1993second}
Babak Hassibi and David~G Stork.
\newblock Second order derivatives for network pruning: Optimal brain surgeon.
\newblock In \emph{Advances in neural information processing systems}, pages
  164--171, 1993.

\bibitem[Helmbold and Long(2017)]{helmbold2017surprising}
David~P Helmbold and Philip~M Long.
\newblock Surprising properties of dropout in deep networks.
\newblock \emph{The Journal of Machine Learning Research}, 18\penalty0
  (1):\penalty0 7284--7311, 2017.

\bibitem[Hinton et~al.(2012)Hinton, Srivastava, Krizhevsky, Sutskever, and
  Salakhutdinov]{hinton_improving_2012}
Geoffrey~E. Hinton, Nitish Srivastava, Alex Krizhevsky, Ilya Sutskever, and
  Ruslan~R. Salakhutdinov.
\newblock Improving neural networks by preventing co-adaptation of feature
  detectors.
\newblock Technical report, July 2012.
\newblock URL \url{http://arxiv.org/abs/1207.0580}.
\newblock arXiv: 1207.0580.

\bibitem[Hyv{\"a}rinen(2013)]{hyvarinen_independent_2013}
Aapo Hyv{\"a}rinen.
\newblock Independent component analysis: recent advances.
\newblock \emph{Phil. Trans. R. Soc. A}, 371\penalty0 (1984):\penalty0
  20110534, February 2013.
\newblock ISSN 1364-503X, 1471-2962.
\newblock \doi{10.1098/rsta.2011.0534}.
\newblock URL
  \url{http://rsta.royalsocietypublishing.org/content/371/1984/20110534}.

\bibitem[Ioffe and Szegedy(2015)]{ioffe2015batch}
Sergey Ioffe and Christian Szegedy.
\newblock Batch normalization: Accelerating deep network training by reducing
  internal covariate shift.
\newblock In \emph{Proceedings of the 32Nd International Conference on
  International Conference on Machine Learning - Volume 37}, ICML'15, pages
  448--456. JMLR.org, 2015.
\newblock URL \url{http://dl.acm.org/citation.cfm?id=3045118.3045167}.

\bibitem[Krizhevsky et~al.(2017)Krizhevsky, Sutskever, and
  Hinton]{krizhevsky_imagenet_2017}
Alex Krizhevsky, Ilya Sutskever, and Geoffrey~E. Hinton.
\newblock {ImageNet} classification with deep convolutional neural networks.
\newblock \emph{Communications of the ACM}, 60\penalty0 (6):\penalty0 84--90,
  May 2017.
\newblock ISSN 00010782.
\newblock \doi{10.1145/3065386}.
\newblock URL \url{http://dl.acm.org/citation.cfm?doid=3098997.3065386}.

\bibitem[Kuncheva and Whitaker(2003)]{kuncheva2003measures}
Ludmila~I Kuncheva and Christopher~J Whitaker.
\newblock Measures of diversity in classifier ensembles and their relationship
  with the ensemble accuracy.
\newblock \emph{Machine learning}, 51\penalty0 (2):\penalty0 181--207, 2003.

\bibitem[LeCun et~al.(1990)LeCun, Denker, and Solla]{lecun1990optimal}
Yann LeCun, John~S Denker, and Sara~A Solla.
\newblock Optimal brain damage.
\newblock In \emph{Advances in neural information processing systems}, pages
  598--605, 1990.

\bibitem[Leisch et~al.(1998)Leisch, Weingessel, and
  Hornik]{leisch1998generation}
Friedrich Leisch, Andreas Weingessel, and Kurt Hornik.
\newblock On the generation of correlated artificial binary data.
\newblock 1998.

\bibitem[Luo(2017)]{luo2017learning}
Ping Luo.
\newblock Learning deep architectures via generalized whitened neural networks.
\newblock In \emph{International Conference on Machine Learning}, pages
  2238--2246, 2017.

\bibitem[Maeda(2014)]{maeda_bayesian_2014}
Shin-ichi Maeda.
\newblock A {Bayesian} encourages dropout.
\newblock Technical report, December 2014.
\newblock URL \url{http://arxiv.org/abs/1412.7003}.
\newblock arXiv: 1412.7003.

\bibitem[Mallat(2012)]{Mallat2012Group}
S.~Mallat.
\newblock Group invariant scattering.
\newblock \emph{Comm. Pure Appl. Math.}, 65\penalty0 (10):\penalty0 1331--1398,
  2012.
\newblock \doi{10.1002/cpa.21413}.
\newblock URL \url{http://dx.doi.org/10.1002/cpa.21413}.

\bibitem[Mariet and Sra(2016)]{mariet2016diversity}
Zelda Mariet and Suvrit Sra.
\newblock Diversity networks.
\newblock 2016.

\bibitem[Peng et~al.(2005)Peng, Long, and Ding]{peng2005feature}
Hanchuan Peng, Fuhui Long, and Chris Ding.
\newblock Feature selection based on mutual information criteria of
  max-dependency, max-relevance, and min-redundancy.
\newblock \emph{IEEE Trans. Pattern Anal. Mach. Intell.}, 27\penalty0
  (8):\penalty0 1226--1238, 2005.

\bibitem[Preisser and Qaqish(2014)]{preisser2014comparison}
John~S Preisser and Bahjat~F Qaqish.
\newblock A comparison of methods for simulating correlated binary variables
  with specified marginal means and correlations.
\newblock \emph{Journal of Statistical Computation and Simulation}, 84\penalty0
  (11):\penalty0 2441--2452, 2014.

\bibitem[Rodr{\'\i}guez et~al.(2016)Rodr{\'\i}guez, Gonzalez, Cucurull,
  Gonfaus, and Roca]{rodriguez2016regularizing}
Pau Rodr{\'\i}guez, Jordi Gonzalez, Guillem Cucurull, Josep~M Gonfaus, and
  Xavier Roca.
\newblock Regularizing cnns with locally constrained decorrelations.
\newblock Technical report, 2016.

\bibitem[Rousseeuw(1987)]{rousseeuw1987silhouettes}
Peter~J Rousseeuw.
\newblock Silhouettes: a graphical aid to the interpretation and validation of
  cluster analysis.
\newblock \emph{Journal of computational and applied mathematics}, 20:\penalty0
  53--65, 1987.

\bibitem[Srivastava et~al.(2014)Srivastava, Hinton, Krizhevsky, Sutskever, and
  Salakhutdinov]{srivastava2014dropout}
Nitish Srivastava, Geoffrey Hinton, Alex Krizhevsky, Ilya Sutskever, and Ruslan
  Salakhutdinov.
\newblock Dropout: a simple way to prevent neural networks from overfitting.
\newblock \emph{The Journal of Machine Learning Research}, 15\penalty0
  (1):\penalty0 1929--1958, 2014.

\bibitem[Tishby and Zaslavsky(2015)]{tishby_deep_2015}
Naftali Tishby and Noga Zaslavsky.
\newblock Deep {Learning} and the {Information} {Bottleneck} {Principle}.
\newblock Technical report, March 2015.
\newblock URL \url{http://arxiv.org/abs/1503.02406}.
\newblock arXiv: 1503.02406.

\bibitem[Tompson et~al.(2014)Tompson, Goroshin, Jain, LeCun, and
  Bregler]{tompson_efficient_2014}
Jonathan Tompson, Ross Goroshin, Arjun Jain, Yann LeCun, and Christopher
  Bregler.
\newblock Efficient {Object} {Localization} {Using} {Convolutional} {Networks}.
\newblock Technical report, November 2014.
\newblock URL \url{http://arxiv.org/abs/1411.4280}.
\newblock arXiv: 1411.4280.

\bibitem[Wager et~al.(2013)Wager, Wang, and Liang]{wager2013dropout}
Stefan Wager, Sida Wang, and Percy~S Liang.
\newblock Dropout training as adaptive regularization.
\newblock In C.J.C. Burges, L.~Bottou, M.~Welling, Z.~Ghahramani, and K.Q.
  Weinberger, editors, \emph{Adv. Neural. Inform. Process Syst.}, volume~26,
  pages 351--359. Curran Associates, Inc., 2013.

\end{thebibliography}
\end{document}